\documentclass{article}

\usepackage{arxiv}

\usepackage[utf8]{inputenc} 
\usepackage[T1]{fontenc}    
\usepackage{hyperref}       
\usepackage{url}            
\usepackage{booktabs}       
\usepackage{amsfonts}       
\usepackage{nicefrac}       
\usepackage{microtype}      
\usepackage{cleveref}       
\usepackage{lipsum}         
\usepackage{graphicx}
\usepackage[numbers]{natbib}
\usepackage{doi}
\usepackage{multirow} 
\title{Robust Subpixel Localization of Diagonal Markers in Large-Scale Navigation via Multi-Layer Screening and Adaptive Matching}

\newif\ifuniqueAffiliation
\uniqueAffiliationtrue

\usepackage{authblk}

\newbox{\orcid}\sbox{\orcid}{\includegraphics[scale=0.06]{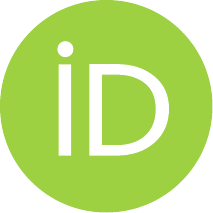}} 
\author[1,2]{Jing Tao}
\author[1,2,*]{Banglei Guan}
\author[1,2,*]{Yang Shang}
\author[1,2]{Shunkun Liang}
\author[1,2]{Qifeng Yu}
\affil[1]{College of Aerospace Science and Engineering, National University of Defense Technology, Hunan 410073, China}
\affil[2]{Hunan Provincial Key Laboratory of Image Measurement and Vision Navigation, Hunan 410073, China}

\renewcommand{\shorttitle}{Applied Optics}

\hypersetup{
pdftitle={A template for the arxiv style},
pdfsubject={q-bio.NC, q-bio.QM},
pdfauthor={David S.~Hippocampus, Elias D.~Striatum},
pdfkeywords={First keyword, Second keyword, More},
}

\begin{document}
\maketitle
\begin{abstract}
	This paper proposes a robust, high-precision positioning methodology to address localization failures arising from complex background interference in large-scale flight navigation and the computational inefficiency inherent in conventional sliding window matching techniques. The proposed methodology employs a three-tiered framework incorporating multi-layer corner screening and adaptive template matching. Firstly, dimensionality is reduced through illumination equalization and structural information extraction. A coarse-to-fine candidate selection strategy minimizes sliding window computational costs, enabling rapid estimation of the marker’s position. Finally, adaptive templates are generated for candidate points, achieving subpixel precision through improved template matching with correlation coefficient extremum fitting. Experimental results demonstrate the method's effectiveness in extracting and localizing diagonal markers in complex, large-scale environments, making it ideal for field-of-view measurement in navigation tasks.
\end{abstract}

\section{Introduction}
		In the field of photogrammetry, cooperative markers have become essential for complex measurement tasks, serving as well-defined geometric benchmarks for manual layout \cite{10972498,high-velocity,AprilTag,10597583}. Unlike naturally occurring features, these artificial markers effectively counteract challenges arising from environmental illumination fluctuations and background texture interference by virtue of their specially designed encoded patterns (\emph{e.g.}, concentric circles \cite{self-similar}, checkerboards \cite{CVPR2011,Structural-Information-Based}, and recent diagonal coding schemes \cite{Liu:25,BullsEye}) and precise geometric constraints. They deliver indispensable advantages in mission-critical applications including but not limited to: autonomous spacecraft rendezvous and docking \cite{solid-state, Landmark-based}, industrial robotic systems requiring high-precision grasping \cite{2018Automatic, HE2025116769}, and vision-based navigation for unmanned aerial vehicles \cite{Multimodal, Robust}. Particularly under large field-of-view conditions involving multi-target measurements, the detection efficiency and subpixel localization accuracy of cooperative markers, especially for low-resolution targets \cite{Planararticle}, directly determine the reliability of subsequent three-dimensional reconstruction and pose estimation processes, thereby constituting a pivotal success factor for photogrammetric mission execution.
		
		\begin{figure*}[t]
			\centering
			\includegraphics[width=0.90\textwidth]{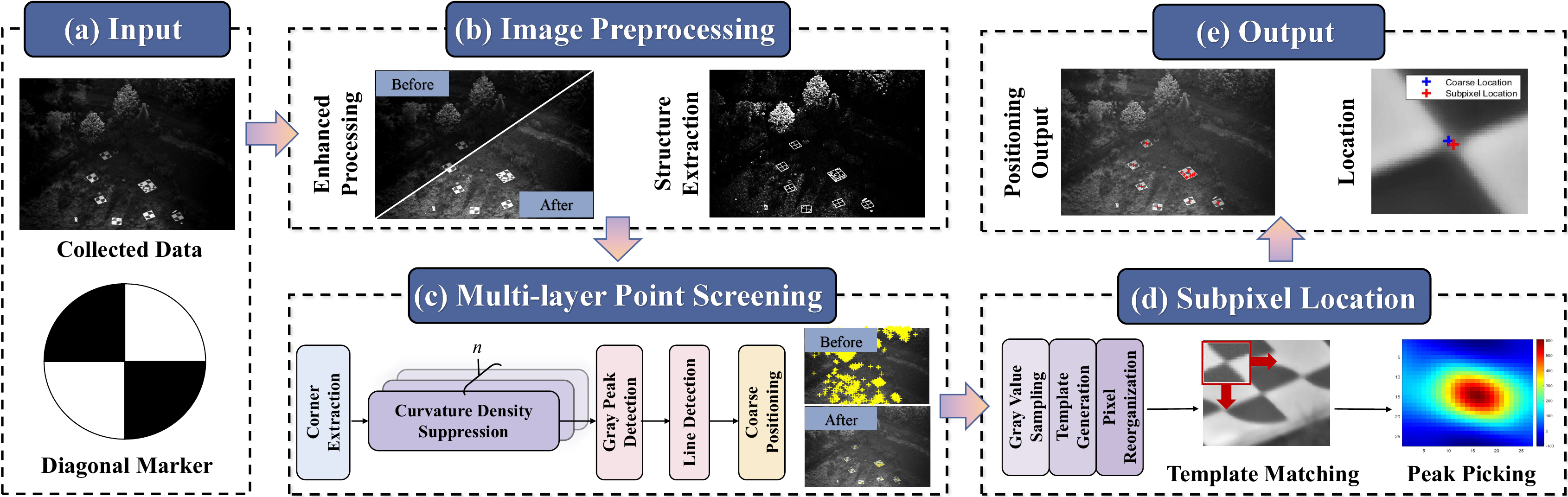}
			\caption{Flowchart of the automated diagonal marker extraction method in complex wide-area scenes.}
			\label{fig:main}
		\end{figure*}
		
		Current research in calibration target detection primarily focuses on circular markers \cite{Dynamic,Fast,2024Circular} and checkerboard patterns \cite{sub-pixel, Event-based, Decoding2022}. Recent advancements have applied deep learning to optimize feature encoding \cite{10079100} and enhance feature robustness \cite{10843758}. Nevertheless, these methods often require extensive training data and exhibit limited interpretability in systems demanding high-precision analysis.
			In contrast, automated extraction methods for diagonally arranged markers have received relatively little attention. Most existing studies concentrate on controlled indoor calibration environments, which may not be directly applicable to complex outdoor settings \cite{2018Automatic, CPIFuse}. Although orthogonal intersecting diagonal markers offer structural simplicity and scalability, their practical application faces significant challenges: 1) In outdoor environments, dynamic illumination changes and background interference can degrade marker features. This necessitates advanced photometric adaptation mechanisms \cite{FENG202450}; 2) For large field-of-view applications where marker regions are only 3–5 pixels wide, existing subpixel methods struggle with background noise suppression.
		
		To enable the efficient and stable extraction of cooperative markers in wide-area vision, this study systematically examines the gradient distribution characteristics of diagonal markers. As shown in Fig. 1, we propose a novel cooperative localization algorithm that integrates a multi-layer corner screening mechanism with adaptive template matching. Experimental results show that the proposed method enhances detection robustness in complex scenarios while maintaining subpixel accuracy. It also provides a feasible technical solution for large-scale photogrammetry. The key contributions of this paper are threefold:
		\begin{itemize}
			\item[$\bullet$] A novel large-scale positioning framework that integrates multi-layer corner screening with adaptive template matching, overcoming traditional localization failures in complex, large-scale environments.
			\item[$\bullet$] An optimized template matching framework that integrates coarse-to-fine hierarchical positioning with advanced matching techniques, overcoming the efficiency bottlenecks typically encountered in conventional two-stage sliding registration methods.
			\item[$\bullet$] A geometry-driven dynamic template generation algorithm that incorporates angular constraints and grayscale distribution modeling. This adaptive system achieves subpixel positioning accuracy under extreme changes in view angles and illumination conditions.
		\end{itemize}
		
		\section{Related Work}
		Subpixel positioning of feature markers is vital in digital photogrammetry as it enables high-precision measurements. In 1973, Hueckel \emph{et al.} introduced the first subpixel edge detection algorithm \cite{1973}, which detects edges at the subpixel level without increasing camera resolution, thereby reducing system costs. Over the years, more advanced subpixel positioning algorithms, such as gradient-based and least-squares fitting methods \cite{9633763,Circular,ZHENG2025102889}, have been developed and widely applied in industrial inspection and remote sensing.
		
		Several methods have also been proposed for diagonal marker extraction. Shang \emph{et al.} \cite{shang} introduced the first automatic recognition framework for circular diagonal markers, providing a new approach to high-precision positioning. Wang \emph{et al.} \cite{2019Precise} proposed the Robust Subpixel Feature Extraction Method (RSFEM), combining template matching and gradient analysis to improve robustness and accuracy in complex environments. Cai \emph{et al.} \cite{2025localization} developed a circular diagonal marker extraction method using the Hough circle detection algorithm for challenging environments. However, these methods still struggle with issues like significant background noise and varying lighting conditions, which compromise accuracy and efficiency. Therefore, developing more effective diagonal marker extraction algorithms remains a key focus.
		
		This paper presents a high-precision visual localization framework that effectively tackles two primary challenges in large-scale navigation: background interference and computational efficiency in sliding window matching. Unlike Ref.~\citenum{2019Precise}, our method abandons template matching in the coarse positioning phase, enabling the detection of more potential candidate points in interference scenarios and greatly improving operational efficiency. We also propose an adaptive template generation method for fine positioning, which enhances the performance of the algorithm. Overall, our framework shows more advantages in computational efficiency and positioning accuracy in large-field-of-view flight scenarios, making it more suitable for high-precision navigation systems on flight platforms.

\section{Method}
		The overall workflow of the algorithm, illustrated in Fig. 1, comprises three core modules: image preprocessing, multi-layer point screening, and subpixel location. The core concept involves first suppressing background interference via image preprocessing to facilitate subsequent extraction. This is followed by coarse positioning to obtain candidate point locations, and finally, fine positioning to achieve subpixel accuracy.
		
		\subsection{Image Preprocessing}
		To mitigate the challenges of non-uniform illumination and noise interference in diagonal marker extraction within complex operational environments, this study introduces a multi-stage collaborative preprocessing framework.
		
		In the initial processing stage, a noise suppression model is constructed using the Gradient-Domain Weighted Guided Image Filter (GDWGIF)\cite{Tao}. The associated energy function is formulated as follows:
		\begin{equation}
			E = \min \sum\limits_{i \in {\Omega _k}} {({{({a_k} \times {I_i} + {b_k} - {q_i})}^2}}  + \frac{\lambda }{{{{\hat T}_I}(k)}} \times {({a_k} - {\psi _k})^2}),
		\end{equation}
		where $q$ denotes the guidance image, $I$ represents the input image, ${a_k, b_k}$ are linear coefficients within local window ${\Omega _k}$, $\hat{T}_I(k)$ constitutes the gradient magnitude-based edge-aware weighting function, $\psi_k$ embodies the gradient domain constraint term, and $\lambda$ is the regularization parameter. 
		
		Distinguished from conventional guided filtering approaches \cite{GIF,wgif}, GDWGIF integrates gradient-domain constraints with adaptive weighting, thereby achieving superior noise attenuation while simultaneously preserving edge integrity and mitigating over-smoothing artifacts. The algorithm details can be found in the author's previous work Ref.~\citenum{Tao}.
		
		\begin{figure*}[t]
			\centering
			\includegraphics[width=0.86\textwidth]{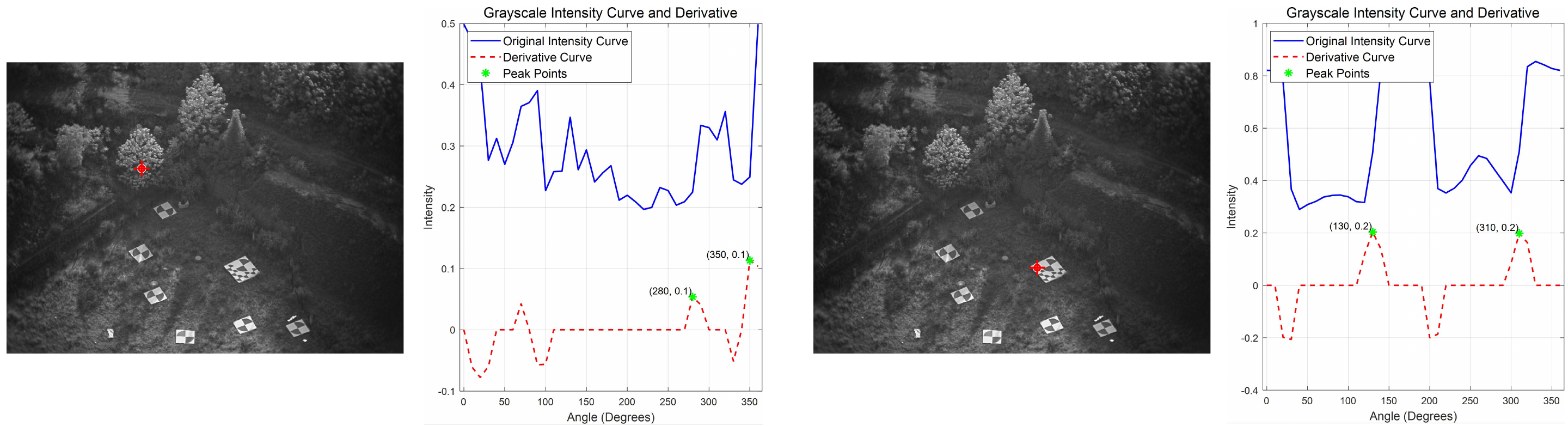}
			\caption{Comparison of annular gradient scanning curves: (a) Grayscale curve of a pseudo-diagonal marker; (b) Grayscale curve of a diagonal marker.}
			\label{fig:circle}
		\end{figure*}
		
		To counteract luminance degradation induced by heterogeneous illumination in field measurement scenarios, a dynamic gamma correction model is implemented for intensity normalization. The model modulates local exposure through a nonlinear mapping between the local neighborhood luminance mean and the gamma parameter:
		\begin{equation}
			{\hat L'_k} = {[{\hat L_k}]^\gamma },\gamma  = \alpha  \cdot \mu (k),
		\end{equation}
		where ${L_k}$ signifies the initial illumination layer, $\mu(k)$ corresponds to the local neighborhood luminance mean, and $\alpha$ functions as a contrast-adaptive scaling factor. 
		
		Unlike conventional global gamma correction, which uses a fixed gamma value and struggles with illumination heterogeneity, our approach offers significant advantages through local adaptability and dynamic parameterization: 1) Gamma values are dynamically adjusted for each region based on local luminance, enabling tailored correction for overexposed and underexposed areas; 2) The contrast-adaptive scaling factor optimizes local contrast during normalization, preventing global contrast imbalance; 3) The model suppresses grayscale saturation in overexposed regions and compression in underexposed regions, reducing luminance distortion while preserving continuity across regional boundaries. Consequently, the model demonstrates superior performance in intensity normalization accuracy and enhanced detail preservation capabilities under complex lighting conditions.
		
		To achieve efficient wide-area feature localization, data dimensionality reduction is achieved through structural information exploitation. As established by Ref.~\citenum{ZHANG2023101895}, structural components encapsulate the predominant information content of images. Based on this, we propose a structural component-based information condensation method that uses multi-scale neighborhood analysis:
		\begin{equation}
			{s_k} = [1 - \exp ( - {(\frac{{{\sigma _d}_{,k}}}{\xi })^t}],
		\end{equation}
		where $d$ indicates the neighborhood diameter across scales, $\sigma_{d,k}$ denotes local standard deviation, and $t$ regulates structural sensitivity. Parameter configurations with $t > 1$ enhance high-frequency edge features, whereas $0 < t < 1$ suppresses noise interference. As demonstrated in Fig. \ref{fig:main}(b), this method adaptively enhances salient structural features through nonlinear contrast mapping, generating a high signal-to-noise ratio feature space conducive to precise diagonal marker localization.
		
		\begin{figure*}[tp]
			\centering
			\includegraphics[width=0.8\textwidth]{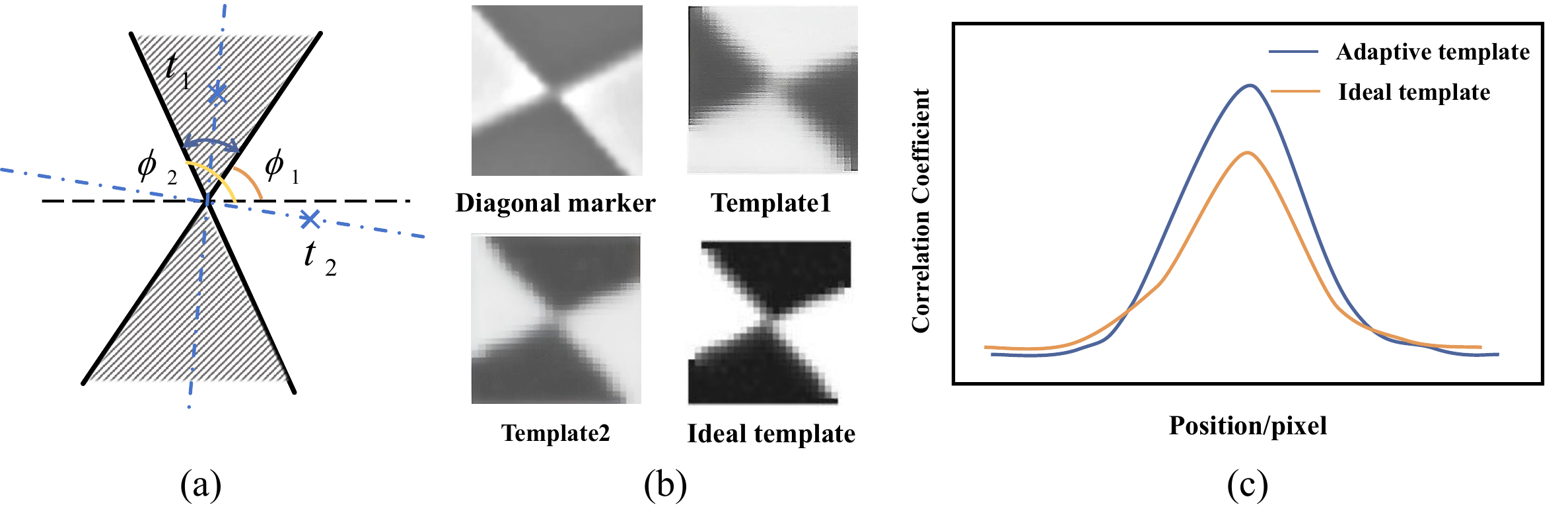}
			\caption{Self-adaptive template generation diagram: (a) Determination of template parameters; (b) Template generation; (c) Correlation coefficient graph.}
			\label{fig:template}
		\end{figure*}
		
		\subsection{Multi-layer Point Screening}
		Given the geometric and imaging characteristics of diagonal markers, we propose a multi-layer point screening framework (see Fig. \ref{fig:main} (c)). This framework first utilizes the prominent corner responses in the central region of diagonal markers in gradient space. A conservative detection strategy based on curvature analysis ensures the completeness of candidate points. By establishing a vector space curvature model (Eq. (\ref{eq:Ck})), we quantify the local curvature features of contour points using the dot product of neighborhood displacement vectors. Normalization ensures rotation-invariant detection. The mathematical expression is as follows:
		\begin{equation}
			\kappa ({p_k}) = 1 - \frac{{\left\langle {{{\overrightarrow v }_{prev}},{{\overrightarrow v }_{next}}} \right\rangle }}{{\left\| {{{\overrightarrow v }_{prev}}} \right\| \cdot \left\| {{{\overrightarrow v }_{next}}} \right\| + \varepsilon }}.
			\label{eq:Ck}
		\end{equation}
		Here, ${\overrightarrow v _{prev}}$ and ${\overrightarrow v _{next}}$ are the displacement vectors within radius $r$ of the current point's front and rear windows, and $\varepsilon $ is a small constant for stability. Matrix operations optimize curvature calculations, enhancing computational efficiency by reducing complexity.
		
		In the primary screening stage, we adopt a dual-threshold mechanism. The peak curvature must exceed three times the standard deviation of the mean, and adjacent peaks must be at least five pixels apart. Combining this with structural component data from preprocessing, we retain only candidate points in the highest 70\% percentile of structural components. Additionally, we propose a noise-aware dynamic region verification criterion. By analyzing connected-domain area distribution and structural noise levels, we establish an adaptive threshold function to effectively eliminate anomalous interference regions. 
		
		Traditional corner detection methods tend to produce excessive feature points in areas with complex textures. To address this issue, we introduce a maximum suppression algorithm based on curvature density analysis. This algorithm combines spatial distribution entropy optimization with adaptive density estimation for intelligent sparse processing of feature points.
		
		To begin with, we construct a curvature-weighted spatial density field, which is mathematically defined as:
		\begin{equation}
			\rho ({p_i}) = \sum\limits_{j \in {\cal N}(i)} \kappa  ({p_j}) \cdot \exp \left( { - \frac{{{{\left\| {{p_i} - {p_j}} \right\|}^2}}}{{2{\sigma ^2}}}} \right),
		\end{equation}
		here, $j$ denotes the index of a neighborhood point around center $i$, and $\kappa ({p_j})$ represents the normalized curvature value. 
		
	   Our approach innovates by integrating a curvature weight coefficient into the density estimation process, unlike the conventional mean shift algorithm \cite{shift}. This allows feature points with higher significance to dominate the density estimation.
		
		During the iterative filtering phase, we apply an improved mean shift strategy to update the positions of candidate points:
		\begin{equation}
			p_i^{new} = \frac{{\sum {{w_j}{p_j}} }}{{\sum {{w_j}} }},{w_j} = \rho ({p_j}) \cdot {e^{ - \left\| {{p_j} - {p_i}} \right\|/h}}.
		\end{equation}
		In this formula, the curvature density weight ${w_j}$ directs feature points towards salient regions, supported by an exponential decay term. The parameter $h$, representing the attenuation coefficient, controls the convergence rate. To enhance spatial distribution, we define spatial grid entropy as:
		\begin{equation}
			H =  - \sum {P(n)\ln P(n)}.
		\end{equation}
		Here, ${P(k)}$ represents the distribution probability of points within the $n$-th grid. By employing a greedy strategy, the algorithm iteratively removes redundant points that yield the maximum entropy increment $\Delta H$ until a predefined entropy threshold is attained. This process effectively reduces spatial overlap while preserving key features, thereby streamlining subsequent candidate point selection.
		
		Diagonal markers maintain two geometric invariants under interference: 1) Center symmetry: Adjacent quadrants exhibit complementary grayscale distributions; 2) Double line intersection: A stable gradient extremum forms along the diagonal direction. To verify these invariants, we design a geometric checker that combines grayscale analysis with LSD line detection \cite{flexible}, implementing a two-stage screening process to ensure structural consistency.
		
		\emph{1) Center symmetry:} This paper introduces a symmetry verification method based on annular gradient scanning (see Fig. \ref{fig:circle}) to address the grayscale distribution of candidate regions. A polar coordinate system is centered on the candidate points, with sampling performed at 10° intervals along an adaptive radius. The grayscale distribution is processed through first-order differentiation and Gaussian difference filtering, transforming extremum detection into zero-crossing identification. This enhances sensitivity to weak edges. The derivative curve of the true diagonal exhibits strict bimodal symmetry (Fig. \ref{fig:circle}(b)), with a phase difference near 180° and opposite gradient polarity, indicating centrosymmetry. In contrast, pseudo-feature points exhibit multi-peak clutter or a dominant single-peak mode (Fig. \ref{fig:circle}(a)). By applying phase difference and gradient polarity constraints, this method reduces reliance on absolute light intensity in traditional grayscale matching, ensuring high verification accuracy even in the presence of noise interference.
		
		\emph{2) Double line intersection:} To efficiently detect straight lines, this study employs the LSD method \cite{flexible}. By verifying intersections of detected lines, the method refines candidate points and determines angle parameters, assisting in the generation of adaptive templates. Specifically, it extracts straight-line features from the local area around the candidate point to check for nearby intersections of two lines. When the detection area contains two straight lines, a line-extension strategy is applied to handle such cases effectively.
		\begin{equation}
			{l_i}^\prime  = Extend({l_i},\Delta s), \Delta s{\rm{ = 5}}.
		\end{equation}
		Here, $\ell  = \{ {l_i}\}$ represents the set of detected straight-line segments, and $Extend()$ extends these segments by $\Delta s$ pixels at both ends. This ensures that potential intersection points are detected, even if the lines are incomplete. To verify the intersection, we compute the Euclidean distance between the intersection of the extended lines and the candidate point. If the distance is below a threshold, the candidate point is retained; otherwise, it is discarded. This process ensures geometric consistency and provides necessary angle parameters for template generation. Experimental results show that this method significantly improves candidate point screening accuracy in complex scenes, offering robust geometric support for template matching.
		
		\subsection{Subpixel Location}
		\subsubsection{Template Adaptive Generation}
		Template matching generally involves feature matching in image space by utilizing a predefined target feature template. However, conventional methods often necessitate the generation of multiple sets of rotated and scaled templates to perform exhaustive matching, which not only leads to significant computational redundancy but also increases the likelihood of mismatches. To tackle these challenges, this study introduces an innovative adaptive template generation framework that is based on the coupling of geometric and photometric information. This approach effectively provides high-precision initial conditions for subpixel positioning, ensuring improved accuracy and efficiency in template matching processes.
		
		\begin{figure}[t]
			\centering
			\includegraphics[width=0.45\textwidth]{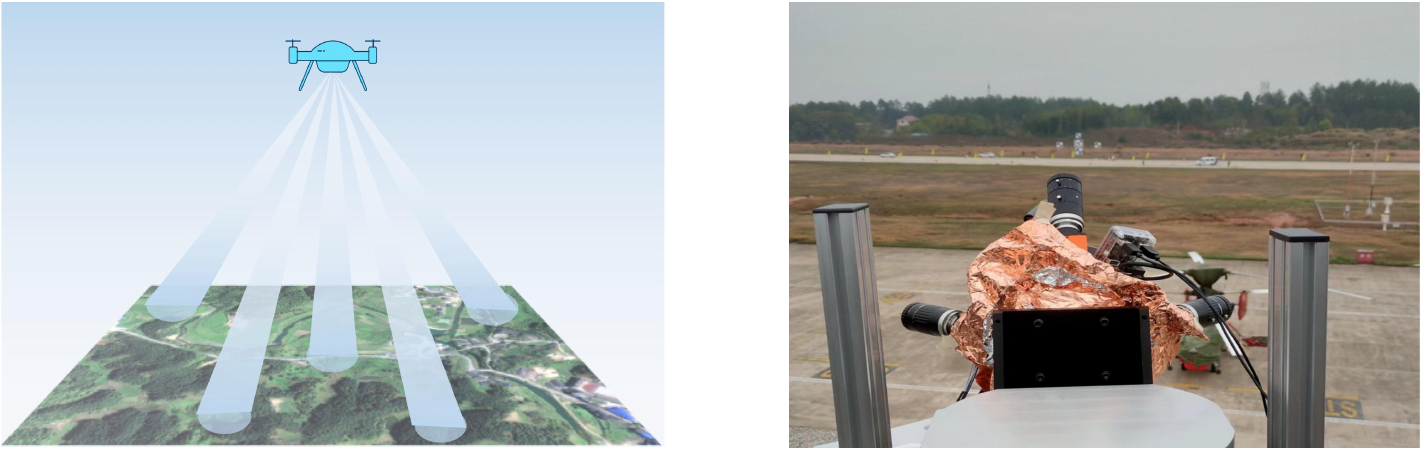}
			\caption{Imaging diagram of the actual project.}
			\label{fig:UAV}
		\end{figure}
		\begin{figure*}[t]
			\centering
			\includegraphics[width=0.75\textwidth]{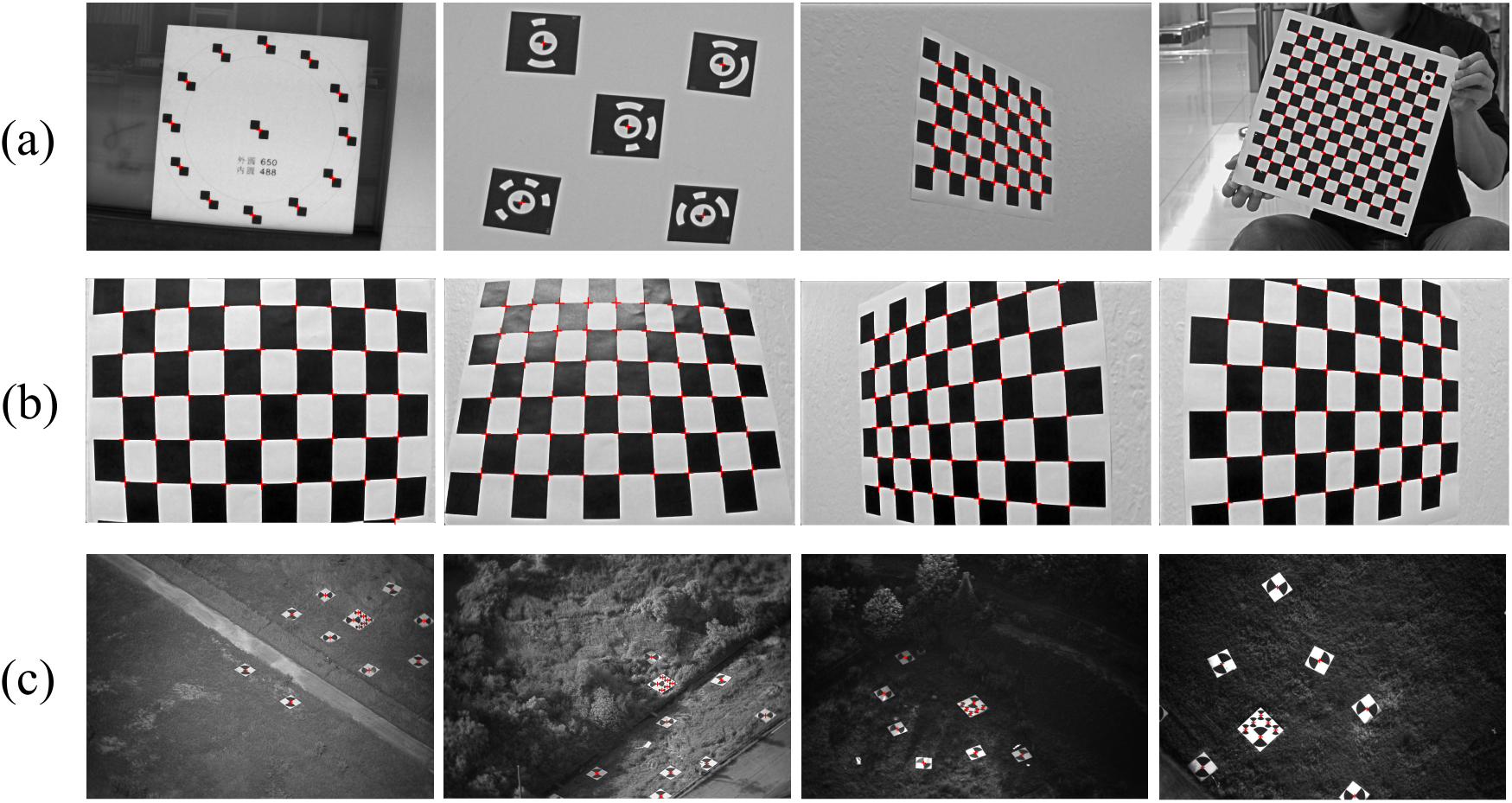}
			\caption{Stability test. (a) Different scenarios; (b) Different perspectives; (c) Complex environment.}
			\label{fig:TEST1}
		\end{figure*}
		
		The core parameters for template generation are derived from geometric and photometric constraints. Using the two straight-line angle parameters, ${\phi _1}$ and ${\phi _2}$, extracted in Section 3.2, we calculate the template angle as $\theta  = {\phi _1} - {\phi _2}$. To model the grayscale distribution, we create a template that simulates the actual diagonal mark. Specifically, within the local coordinate system, we design a $3 \times 3$ sampling window, offset by 5 pixels along the direction angle $\theta /2$ (illustrated as the blue cross in Fig. \ref{fig:template}(a)). To better align the template with real-world imaging conditions, we apply a defocus degradation model based on the optical system. Using the blur parameter ${\sigma _b}$, estimated from the Point Spread Function (PSF), we perform Gaussian convolution on the template to simulate defocus blur. This frequency-domain convolution ensures that the template's high-frequency features are spectrally consistent with the actual image.
		
		To resolve the matching ambiguity caused by the rotational symmetry of the template (as shown in the comparison between template 1 and template 2 in Fig. \ref{fig:template}(b)), we introduce a criterion based on feature space projection:
		\begin{equation}
			{m^*} = \arg {\min _{m \in \{ 1,2\} }}{\left\| {{\rm{P - }}{{\rm{T}}^{(m)}}} \right\|_1},
		\end{equation}
		where ${\rm{P}}$ represents the local image block sampling vector, and ${{\rm{T}}^{(m)}}$ denotes the sampling vector of the two possible templates. To assess the effectiveness of the generated templates, the correlation coefficient curves for the two templates are shown in Fig. \ref{fig:template}(c). The results indicate that the adaptive template exhibits a more ideal single-peak characteristic, improving the accuracy of subsequent subpixel positioning.
		
		This adaptive template generation framework not only significantly reduces computational complexity but also enhances the accuracy and robustness of template matching, offering reliable support for subpixel positioning in complex scenes.
		
		\subsubsection{Subpixel Positioning Techniques}
		Subpixel target localization represents a significant advancement in enhancing the precision of cooperative marker measurement systems, offering substantial theoretical and practical engineering value. After the initial coarse positioning of feature points, further refinement of their coordinates within the localized extraction region is essential to achieve subpixel accuracy. Traditional methods that employ Normalized Cross-Correlation (NCC) template matching are computationally constrained due to their inherent mechanisms. Specifically, the sliding-window approach necessitates pixel-wise traversal to generate correlation coefficient matrices, resulting in prohibitive computational complexity that renders real-time implementation impractical. Drawing on the findings of Ref.~\citenum{Log-Gabor}, this study introduces an accelerated matching framework. By integrating feature recombination, matrix operations, and quadratic surface extremum fitting, this framework achieves efficient subpixel positioning.
		
		The core acceleration strategy reconfigures the neighborhood mapping operation through tensor reshaping. Given a template window of size $(2r + 1) \times (2r + 1)$ and an input feature map $F \in {{\rm{R}}^{H \times W \times d}}$, we construct the expanded feature tensor:
		\begin{equation}
			\tilde F(x,y,(z - 1){r^2} + (i - 1)r + j) = F(x + i - r,y + j - r,z),
		\end{equation}
		where $\tilde F \in {{\rm{R}}^{H \times W \times ({r^2}d)}}$, $(H,W,d)$ represent the spatial dimensions and feature dimension, $(i,j)$ are the template coordinates, and $z$ is the feature dimension index. The NCC calculation between the reference template ${\tilde F_{ref}}$ and the target region ${\tilde F_{tar}}$ is then reformulated as matrix multiplication by reshaping the expanded tensor:
		\begin{equation}
			R = \frac{{{{\tilde F}_{ref}} \cdot {{\tilde F}_{tar}}}}{{{{\left\| {{{\tilde F}_{ref}}} \right\|}_2}{{\left\| {{{\tilde F}_{tar}}} \right\|}_2}}}.
		\end{equation}
		
		This transformation reduces the time complexity from $O(MN{k^2})$ to $O(MN + {k^2})$, significantly improving computational efficiency.
		
		After obtaining the correlation coefficient matrix $R$, a local neighborhood quadratic surface model is constructed:
		\begin{equation}
			c(x,y) = a{x^2} + b{y^2} + cxy + dx + ey + f,
		\end{equation}
		where the weighted least squares method is applied to solve for the coefficients $\{ a,b,c,d,e,f\} $, and the subpixel offset is determined as follows:
		\begin{equation}
			\Delta x = \frac{{2bc - ce}}{{{c^2} - 4ab}},\Delta y = \frac{{2ae - cd}}{{{c^2} - 4ab}}.
		\end{equation}
		By applying these offsets to the coarse positioning results, high-precision target position correction is achieved, completing the subpixel localization. The experimental results, shown in Fig. \ref{fig:main}(e), demonstrate that the red dot position is significantly more accurate than the blue dot position.
		
		\begin{table*}[t]
			\centering
			\caption{Deviations of checkerboard corner detection by two methods}
			\renewcommand\arraystretch{1.2} 
			\resizebox{0.7\textwidth}{!}{ 
				\begin{tabular}{cccc}
					\toprule
					\multicolumn{1}{c}{\multirow{2}[2]{*}{Image number}} & \multicolumn{1}{c}{\multirow{2}[2]{*}{Maximum deviations/pixel}} & \multicolumn{1}{c}{\multirow{2}[2]{*}{Average deviations/pixel}} & \multicolumn{1}{c}{\multirow{2}[2]{*}{Deviations $\leq 0.1$ pixel/\%}} \\
					&       &       &  \\
					\midrule
					1     & 0.2512 & 0.0181 & 98.57 \\
					2     & 0.128 & 0.0146 & 99.47 \\
					3     & 0.2094 & 0.0185 & 98.04 \\
					4     & 0.3457 & 0.0211 & 97.93 \\
					5     & 0.1582 & 0.0151 & 98.74 \\
					6     & 0.0924 & 0.01  & 100 \\
					\bottomrule
				\end{tabular}%
				\label{tab:table1}}
		\end{table*}%

\section{Experiment}
		This study evaluates the proposed diagonal marker extraction algorithm using both a laboratory-constructed multi-form diagonal sign dataset and real-world engineering measurement data. To ensure scientific rigor and comparability, we benchmark the performance of the proposed algorithm against Zhang's checkerboard point extraction algorithm \cite{flexible}, which is widely recognized for its precision and broad applicability, particularly in camera calibration tasks. This algorithm serves as an ideal comparison for validating the accuracy of the proposed method.
		
		\begin{figure}[t]
			\centering
			\includegraphics[width=0.45\textwidth]{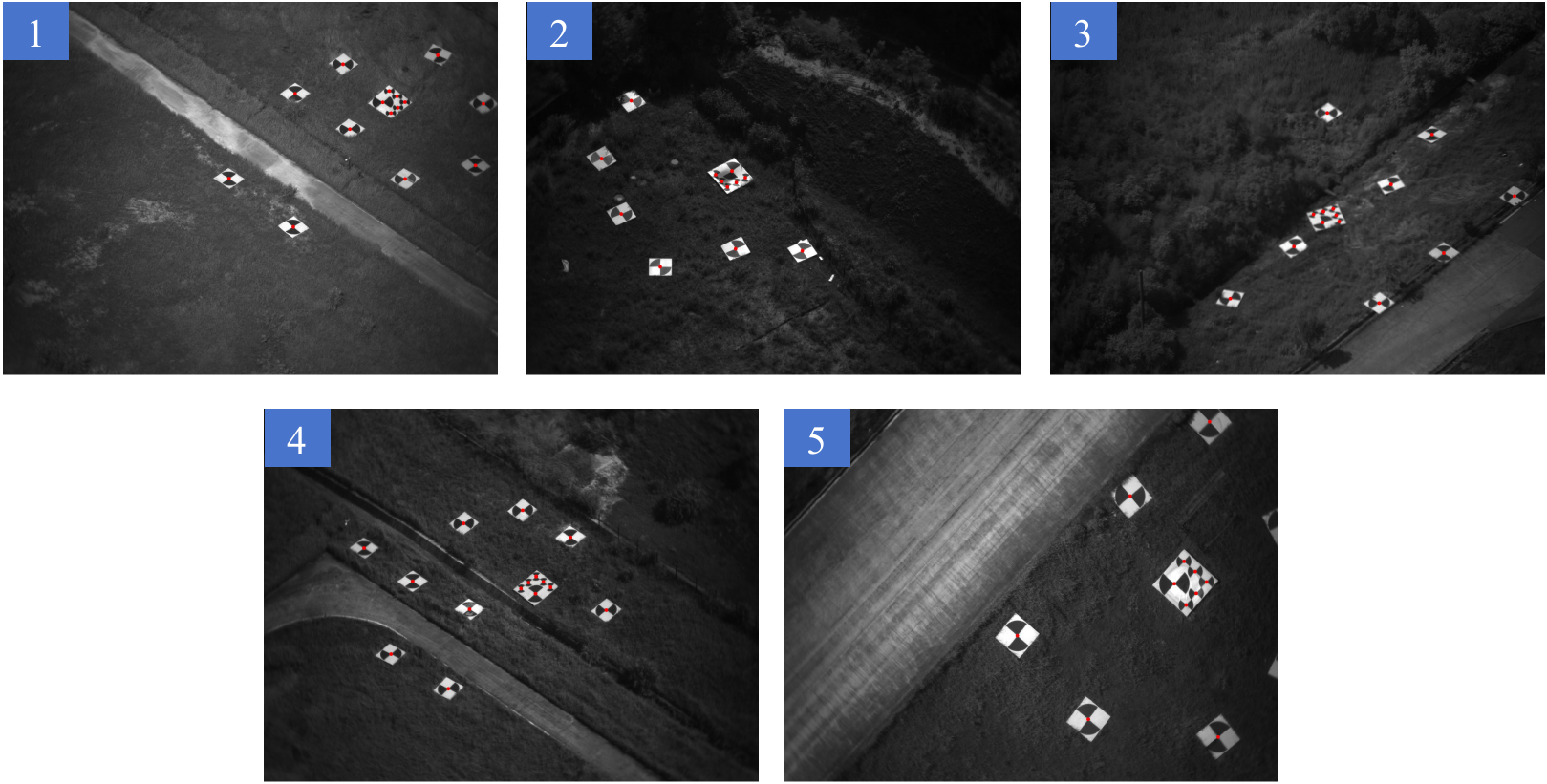}
			\caption{The sample images obtained from the flight experiment and the results of marker extraction.}
			\label{fig:flight}
		\end{figure}
		
		In addition to the laboratory dataset, we further assess the algorithm's accuracy and robustness using engineering measurement data obtained from a multi-aperture flight platform. Fig. \ref{fig:UAV} illustrates the engineering scene captured by the platform. The data acquisition process involves an Unmanned Air Vehicle (UAV) equipped with a multi-aperture imaging system, which captures aerial images of a ground-based cooperative target. The system features a single camera with a resolution of 4096 × 3000 pixels, paired with a 150 mm fixed-focus lens. The algorithm is implemented in the Matlab R2021a development environment.
		
		\subsection{Robustness Verification}
		To validate the algorithm's robustness in complex scenarios, we conducted comprehensive experiments across three critical dimensions: adaptability to multi-form cooperative markers, stability under perspective variations, and performance in challenging field environments. Fig. \ref{fig:TEST1} illustrates the feature extraction performance under various test conditions.
		
		Fig. \ref{fig:TEST1} (a) shows the detection results for multi-scenario cooperative markers, including standard checkerboard patterns and specially designed diagonal markers. The algorithm effectively handles diverse structural configurations by leveraging geometric constraints of diagonal features. Fig. \ref{fig:TEST1} (b) demonstrates the algorithm's stable positioning accuracy under varying observation angles, despite target deformation and scale changes caused by perspective projection. Fig. \ref{fig:TEST1} (c) highlights the system's performance in outdoor environments, successfully addressing challenges such as wide-field imaging, illumination variations, background interference, and motion blur.
		
		The experimental results show that the proposed algorithm achieves robust multi-scenario extraction by simply inputting the target’s initial physical dimensions, eliminating the need for manual parameter tuning. It shows superior adaptability to varying geometries, wide angles, and dynamic interference, proving its practical value in engineering. In contrast, mainstream methods primarily focus on checkerboard pattern detection, with limited capability for diagonal marker localization in complex environments. For instance, Zhang’s checkerboard corner detection algorithm\cite{flexible} and existing diagonal marker methods\cite{2019Precise,2025localization} fail to reliably position diagonal markers in the challenging scenarios shown in Fig. \ref{fig:TEST1} (c).
		
		\begin{figure*}[t]
			\centering
			\includegraphics[width=0.8\textwidth]{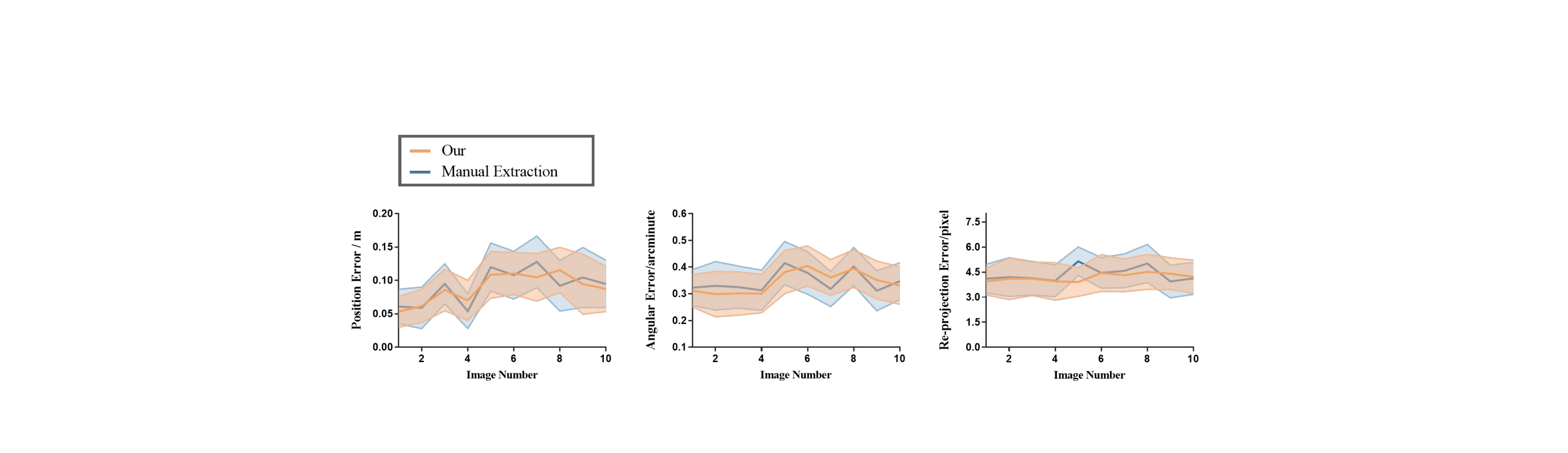}
			\vspace{0.2cm}
			\caption{Measurement error results of the two methods in flight experiments.}
			\label{fig:DATA}
		\end{figure*}
		\begin{figure*}[t]
			\centering
			\includegraphics[width=0.7\textwidth]{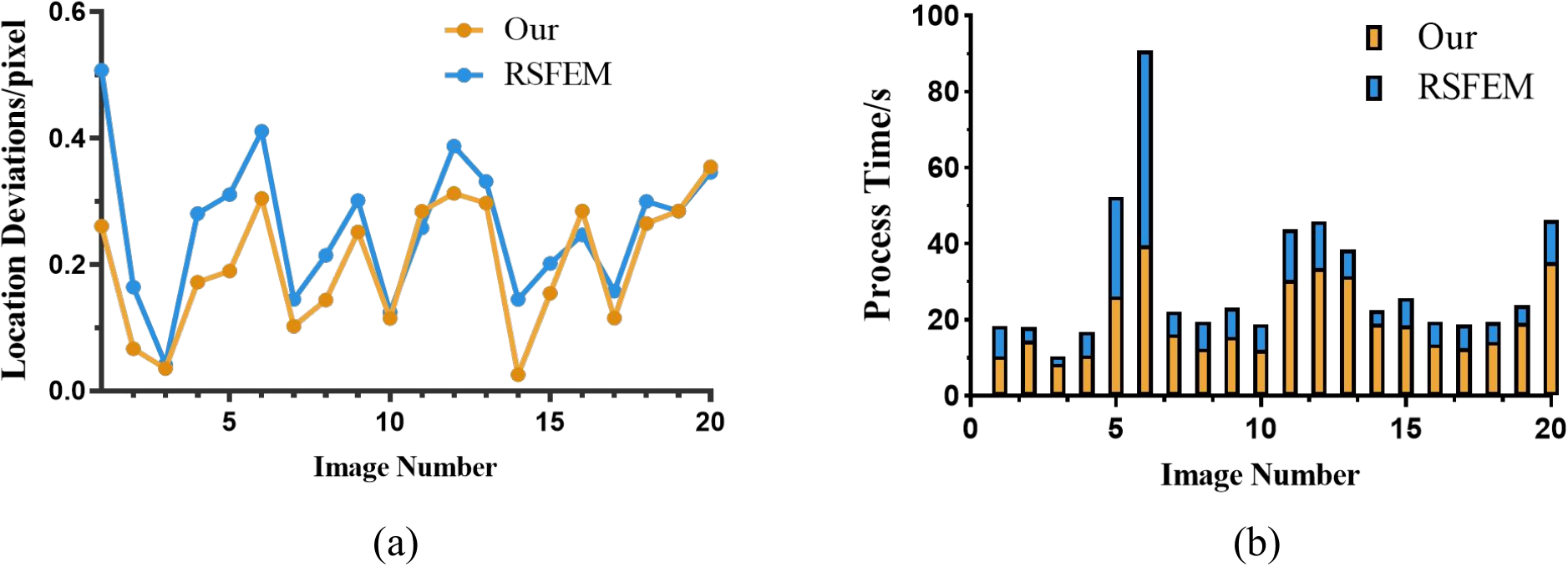}
			\vspace{0.2cm}
			\caption{Comparison of the results of the two methods under flight experiments: (a) positioning errors; (b) algorithm processing time.}
			\label{fig:time}
		\end{figure*}
		
		\subsection{Precision Verification}
		\subsubsection{Checkerboard Positioning}
		To validate the positioning accuracy of the proposed algorithm, we conducted comparative experiments under controlled laboratory conditions using identical checkerboard patterns. As shown in Fig. \ref{fig:TEST1} (b), both our method and Zhang's conventional algorithm \cite{flexible} were applied to extract corner coordinates from standardized calibration images. The quantitative evaluation was performed by analyzing the differences in detected corner positions between the two methods.
		
		As detailed in Table \ref{tab:table1}, the maximum deviations observed for both methods approach 0.3 pixels, which can be attributed to angular distortions of the checkerboard pattern, especially at the edge corner points. However, overall, the proposed method demonstrates an average deviation of less than 0.03 pixels, with over 97\% of corner points deviating by less than 0.1 pixels. These results show that the proposed methodology provides positioning precision comparable to Zhang's established corner extraction approach \cite{flexible} in controlled environments, meeting the stringent accuracy requirement ($\leq 0.1$ pixel) for practical engineering applications.
		
		\subsubsection{Flight Platforms Positioning}
	    As depicted in fig. \ref{fig:flight}, the multi-aperture aerial positioning system employs five synchronized long-focus optical sensors to capture images of ground-based cooperative targets, achieving spatial positioning for the flight platform through an absolute pose estimation algorithm. To evaluate the performance of the proposed adaptive positioning algorithm under dynamic flight conditions, a quantitative comparison of positioning errors was conducted among three methods: manual feature extraction \cite{shang}, the proposed automatic feature extraction method, and reference measurements from the Position and Orientation System (POS).
		
		Fig. \ref{fig:DATA} shows the accuracy assessment of our feature extraction algorithm using real-flight imagery. Positional precision is evaluated by the Euclidean distance between the derived coordinates and ground-truth positions. For directional accuracy (specifically azimuth/orientation), the angular error is calculated as follows: First, the re-projection error (in pixels) is determined. This pixel error is then converted to an angular error (in radians) by multiplying it with the camera's angular resolution, which is intrinsically linked to the calibrated focal length of the optics. Finally, the angular error is typically reported in arc-minutes for clarity. The detailed mathematical derivation and specific steps of this angular error calculation process can be found in Ref~\citenum{liang2024accurate}. The error envelope and curve in Fig. \ref{fig:DATA} demonstrate that our automatic feature extraction method matches the accuracy of manual extraction, exhibiting a slightly narrower fluctuation range which indicates improved robustness. These results confirm that the proposed positioning algorithm achieves centimeter-level spatial accuracy and arc-minute-level orientation accuracy, proving its effectiveness in dynamic aerial environments.
		
		To comprehensively evaluate the robustness of the algorithm, this study compares the proposed method with a previously developed automatic corner extraction algorithm RSFEM \cite{2019Precise}. The comparison reveals that the template matching method in the RSFEM's coarse positioning process performs poorly when the flight platform's imaging quality is subpar. To ensure a fair and feasible comparison, we incorporate local region of interest (ROI) segmentation and coarse position guidance into the process. As depicted in Fig. \ref{fig:time} (a), the error distribution profiles indicate that our method consistently achieves subpixel precision, even in challenging field conditions. Quantitative analysis reveals significant improvements: approximately a 29\% reduction in mean coordinate deviation and a 15\% decrease in dispersion metrics. These enhancements are attributed to our scene-adaptive template generation framework, which dynamically adjusts the matching mode based on environmental feedback. This innovative approach effectively reduces the sensitivity of traditional algorithms to background clutter, significantly enhancing the positioning system's reliability in complex environments.
		
		\subsection{Computational Efficiency}
		To assess the real-time processing capabilities, we conducted comparative benchmarking of computational efficiency across different methods using standardized hardware configurations. As shown in Fig. \ref{fig:time} (b), our method demonstrates faster processing throughput than RSFEM \cite{2019Precise} in large-scale scenarios (4096×3000 resolution). Notably, while scene complexity significantly affects the efficiency of diagonal marker localization in traditional approaches, our method maintains less than 12\% variance in efficiency across test sequences. These results underscore the superior computational stability of our method, with processing time deviations 38\% lower than those of the comparison methods. Future work will focus on implementing CUDA-accelerated parallel computing architectures and CPU SIMD optimization techniques to further enhance real-time performance in task-critical applications.
		
		\section{Conclusion}
		This study tackles two key challenges in wide-area aerial navigation: positioning failures caused by complex background interference and the inefficiency of traditional sliding-window matching in large-scale scenarios. We propose a diagonal marker positioning method that combines multi-layer corner screening and adaptive template matching in a three-tiered processing framework. First, data is simplified through illumination equalization and structural information extraction. Next, a coarse-to-fine candidate point screening strategy reduces computational overhead and efficiently estimates the rough marker position. Finally, adaptive template generation and an improved matching algorithm enable subpixel-level positioning using the fitting correlation coefficient extremum method. Experimental results demonstrate the stability of the proposed method in complex, wide-area scenes. Our algorithm outperforms existing methods in accuracy and robustness, effectively addressing positioning challenges in dynamic flight platforms under strong interference. This research not only advances feature extraction theory in complex scenes but also lays a foundation for UAV autonomous navigation and aerial surveying applications.
		
		In future work, we aim to further enhance the applicability and real-time performance of the algorithm, with the goal of achieving versatile performance across multiple scenarios. Additionally, we plan to integrate deep learning algorithms to further improve the robustness of feature extraction.

		\section*{Acknowledgement}
		This work was supported by the Hunan Provincial Natural Science Foundation for Excellent Young Scholars (Grant No. 2023JJ20045), the Foundation of National Key Laboratory of Human Factors Engineering (Grant No. GJSD22006) , and the National Natural Science Foundation of China (Grant No. 12372189).
		
		\section*{Disclosures}
		The authors declare no conflicts of interest.
		
		\section*{Data Availability}
		Data underlying the results presented in this paper are not publicly available at this time, but may be obtained from the authors upon reasonable request.

\end{document}